\documentclass[letterpaper, 10 pt, conference]{ieeeconf}  

\IEEEoverridecommandlockouts                              

\usepackage{comment}
\usepackage{graphicx} 
\usepackage{amsmath} 
\usepackage{amssymb}  
\usepackage{subcaption}
\usepackage[dvipsnames]{xcolor}
\usepackage{multirow}
\usepackage{placeins}
\usepackage{balance}
\usepackage{url}
\usepackage{hyperref}

\title{\LARGE \bf ContactGrasp: Functional Multi-finger Grasp Synthesis from Contact}

\author{
	Samarth Brahmbhatt$^{1,2}$,
	Ankur Handa$^{2}$,
	James Hays$^{1}$,
	and Dieter Fox$^{2}$
	\thanks{
		$^{1}$Institute of Robotics and Intelligent Machines, Georgia Tech, Atlanta, GA 30332, USA
	}%
	\thanks{
		$^{2}$NVIDIA Robotics, Seattle, WA 98105, USA
	}%
}

\begin{document}

\makeatletter
\let\@oldmaketitle\@maketitle
\renewcommand{\@maketitle}{\@oldmaketitle
  \includegraphics[width=\linewidth]{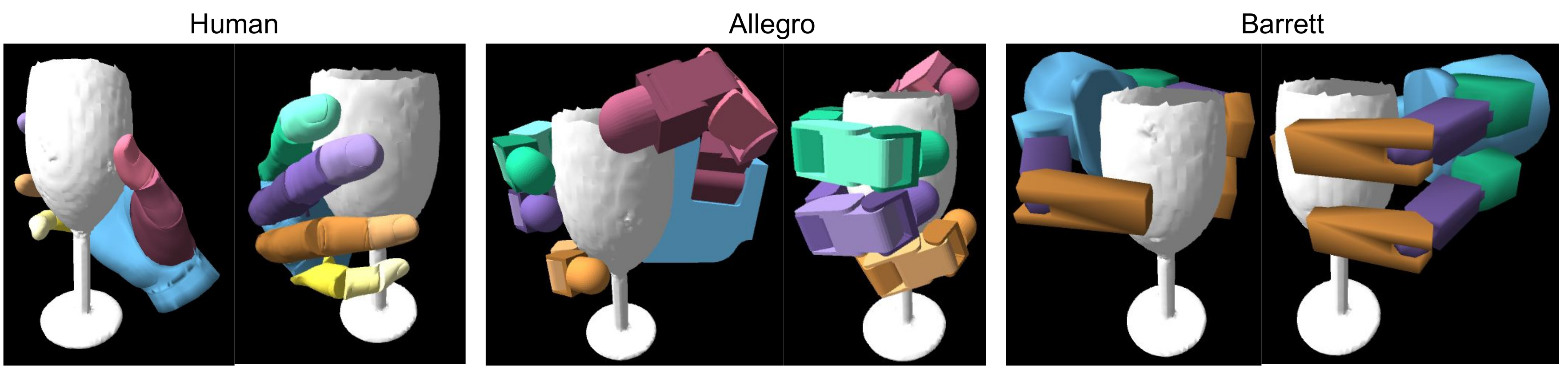} \\[0.35em]
  \refstepcounter{figure}\footnotesize{Fig.~\thefigure. ContactGrasp synthesizes functional grasps for diverse hand models. Here we show grasps for `using' a wine glass to drink from it. Left: 20-DOF human hand, Middle: 16-DOF Allegro hand, and Right: 4-DOF Barrett hand. Each hand has additional 6 DOFs for palm pose. }
  \label{fig:teaser} \medskip \vspace{-10pt}}
\makeatother

\maketitle
\thispagestyle{empty}
\pagestyle{empty}

%

\begin{abstract}
Grasping and manipulating objects is an important human skill. Since most objects are designed to be manipulated by human hands, anthropomorphic hands can enable richer human-robot interaction. Desirable grasps are not only stable, but also functional: they enable post-grasp actions with the object. However, functional grasp synthesis for high degree-of-freedom anthropomorphic hands from object shape alone is challenging. We present ContactGrasp, a framework for functional grasp synthesis from object shape and contact on the object surface. Contact can be manually specified or obtained through demonstrations. Our contact representation is object-centric and allows functional grasp synthesis even for hand models different than the one used for demonstration. Using a dataset of contact demonstrations from humans grasping diverse household objects, we synthesize functional grasps for three hand models and two functional intents. The project webpage is \url{https://contactdb.cc.gatech.edu/contactgrasp.html}.
\end{abstract}

\section{Introduction}\label{sec:introduction}
Household objects are designed for use and manipulation by human hands. Humans excel at grasping and then performing actions with these objects. Enabling this skill for robots has the potential to unlock more productive and natural human-robot interaction. We take a step towards this by proposing ContactGrasp, a framework to synthesize functional grasps. Using object shape and contact demonstrations, ContactGrasp allows functional grasp synthesis for kinematically diverse hand models.


Recent work on robotic grasping of household objects focuses on using large amounts of data collected by trying random grasping actions, often using parallel-jaw or suction-cup end effectors~\cite{mahler2017dex, mahler2017suction, mahler2019learning, pinto2016supersizing}. This approach generates lots of self-supervised data that enables training of robust grasp policies. However, the simplicity of the end effectors has a big hand in enabling this data-collection strategy. In addition, such end effectors are mostly suited for pick-and-place tasks and do not allow performing a post-grasp action (e.g. handing an object off, clicking the camera shutter button, switching on a flashlight, etc.). Functionality of a grasp is important because 1) it enables more natural collaboration in activities with humans, and 2) most household objects are designed with specific functional grasps in mind (e.g. spray bottle has contoured neck and squeezer, hammer has a long handle, etc.).

A large body of work addresses grasp synthesis from object geometry. These approaches model the hand as a kinematic tree of rigid mesh parts and intelligently sample its configuration space to synthesize a set of stable grasps~\cite{ciocarlie2007dimensionality, roa2012power, przybylski2011planning}. However, these grasps lack the notion of functionality and most of them are far from how a human would grasp the object (see Figure~\ref{fig:overview} for examples).

Other approaches have utilized human demonstrations to close this gap. The human hand pose (captured by data gloves in~\cite{ekvall2007learning}, and by visual recognition in~\cite{romero2009modeling}) is mapped by hand-engineered transformations to a similar robot end-effector pose. Kinesthetic teaching~\cite{herzog2012template} can deliver demonstrations directly in the target hand model space. However, such kinematic re-targeting methods are tied to a specific end-effector and require careful analysis to develop mappings to new end-effectors. Approaches that record the human hand pose suffer from the additional difficulty of orienting the hand pose w.r.t. the object (which requires  embedding an additional 6-DOF magnetic tracker in the object~\cite{Garcia-Hernando_2018_CVPR}). In addition, they lack a clear objective to reproduce the demonstrated contact, as we show in Section~\ref{sec:results}. Analysis of contact during human grasping has shown that humans prefer to contact specific areas during functional grasping~\cite{contactdb}.

This motivates the development of an object-centric approach that is not tied to a specific end-effector, and that emphasizes the reproduction of demonstrated contact. Towards this end, we propose ContactGrasp, a framework which synthesizes grasps from both object geometry and contact on the object surface. Contact can be specified manually, through human demonstrations (e.g. ContactDB~\cite{contactdb}), or a combination of both. We show in Section~\ref{sec:results} that ContactGrasp can be used to synthesize grasps that reproduce the demonstrated contact for multiple different hand models. Grasp synthesis from contact has the drawback that multiple hand configurations can sometimes result in the same contact pattern. To address this, ContactGrasp adopts a sample-and-rank approach which outputs a ranked set of grasps. We show qualitatively and quantitatively in Section~\ref{sec:results} that the desired grasp can be found among the top ranked grasps in this set.

To summarize, we make the following contributions in this paper:
\begin{itemize}
	\item Develop a multi-point \textbf{contact representation} that supports efficient grasp synthesis.
	\item Propose a sample-and-rank approach for \textbf{functional grasp synthesis} from object shape \textit{and contact} that can work with multiple hand models. 
\end{itemize}
\section{Related Work} \label{sec:related_work}
Grasp synthesis has been widely studied from many perspectives. \textit{Analytic} approaches~\cite{bicchi2000robotic, prattichizzo2012manipulability, rosales2012synthesis, nguyen1988constructing, roa2009computation, krug2010efficient, rodriguez2012caging, seo2012planar} synthesize grasp(s) from object shape, which usually guarantee stability within their set of assumptions. These assumptions include perfect object models, rigid body approximation of the hand, Coulomb friction, simplified contact models, etc. While such approaches contribute valuable theoretical analysis of grasping, their simplifying assumptions are often quite restrictive. This makes purely analytic algorithms difficult to deploy in the real world.

In contrast, \textit{data-driven} approaches rely on machine learning techniques to learn features in some representation of the object (RGB image, 3D mesh, etc.) that can be used to predict grasps. For example, Li et al~\cite{li2005shape} compute shape features for both objects and grasping hands, and use nearest-neighbor retrieval to synthesize grasps for novel objects. Newer deep-learning based algorithms typically require large amounts of training data, which is expensive to collect and label. In addition, they are limited to simple end effectors like parallel jaw grippers to enable efficient training data collection. For example,~\cite{lenz2015deep, jiang2011efficient, DBLP:journals/corr/RedmonA14} predict grasps for a parallel jaw gripper from a dataset of manually annotated images. Some recent works like Pinto et al~\cite{pinto2016supersizing} have used self-supervision to automate data collection, again for a simple parallel jaw gripper. We refer the reader to~\cite{bohg2014data} for an extensive survey of data-driven grasp synthesis.

\textit{Hybrid} approaches use analytic criteria (e.g.~\cite{miller1999examples, ferrari1992planning}) to sample a large number of grasps in a simulator like GraspIt!~\cite{miller2004graspit}. Real-world demonstrations in various forms are then used to process these samples: filter them or train a machine learning algorithm to predict success. For example, Mahler et al~\cite{mahler2017dex, mahler2019learning, mahler2017suction, pinto2016supersizing} execute those grasps with a robot and record success/failure. Song et al~\cite{song2011multivariate} learn a Bayes Net to jointly model post-grasp task and discretized hand pose. Synthetic data generated using a grasp planner is labeled manually for post-grasp task suitability. However, the algorithm is not aware of contact and the pose discretization often results in predictions that are not in contact with the object.

\textbf{Grasp Synthesis from Contact}: Hamer et al~\cite{hamer2010object} record human demonstrations of grasping by in-hand scanning to get both object and hand pose. Contact points are aggregated on the object surface and used to form a prior for hand pose synthesis and tracking. In contrast to ContactGrasp, their algorithm requires demonstrations of both hand pose and contact. Ben Amor et al~\cite{amor2012generalization} learn a low-dimensional grasp space from human demonstrations acquired using a data-glove. This is used along with manually specified contact points to optimize the final robot grasp. In addition to requiring manual specification of per-finger contact point, it is not clear how well the low-dimensional space can recover the finegrained functional grasps synthesized by ContactGrasp. Varley et al~\cite{varley2015generating} replace human demonstrations with synthetic data by running a grasp planner in simulation and recording the fingertip contact points. Ye et al~\cite{ye2012synthesis} develop an algorithm to sample physically plausible contact points between the hand the object, given the wrist and object pose from a motion-capture system. These sampled points are then used to synthesize realistic-looking hand poses. These methods approximate contact as a single point per fingertip. In contrast, ContactGrasp allows realistic multi-point contact (see Figure~\ref{fig:contact_map}), allows using kinematically diverse hand models, and demonstrates good performance across more complex and numerous objects.
\section{Contact Model and Human Demonstrations} \label{sec:contact}
As mentioned in Section~\ref{sec:introduction}, ContactGrasp can leverage human demonstrations of grasp contact to synthesize similar grasps for various hand models. It has been shown that humans contact grasped objects with not only fingertips, but also the palm and non-tip areas of the fingers. Hence our contact model and grasp synthesis algorithm supports multi-point contact.

\begin{figure}
    	\centering
    	\includegraphics[width=0.35\textwidth]{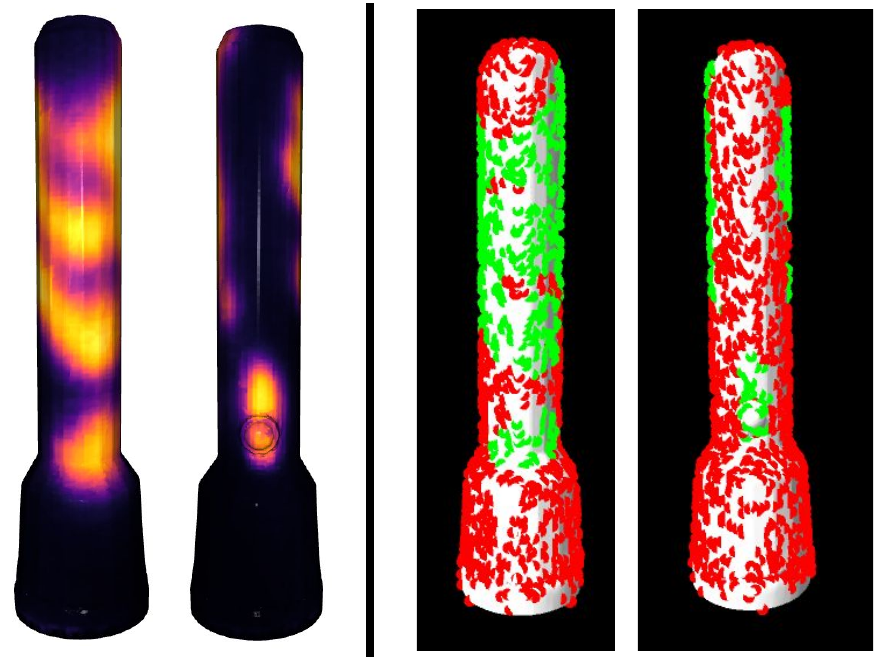}
    \caption{Contact map construction for the `flashlight' object from ContactDB human demonstration. Points are randomly sampled on the object surface. \textcolor{OliveGreen}{green}: attractive, \textcolor{red}{red}: repulsive.}
    \label{fig:contact_map}
\end{figure}

We define the contact map $\mathbf{c}$ as a set of $N$ points $p_i$ sampled uniformly at random on the object surface, with a contact value $c_i$ of $+1$ (attractive) or $-1$ (repulsive). Contacted points in the demonstration are marked attractive, while others are marked repulsive. Repulsive points allow ContactGrasp to exploit negative information~\cite{schmidt2014dart}. Figure~\ref{fig:contact_map} shows an example of the contact map for the `flashlight' object, with attractive points in green and repulsive points in red.

\subsection{Human Contact Demonstrations}
The contact model specified above supports manual specification. However, most of our experiments are performed using real-world contact maps from the ContactDB dataset~\cite{contactdb}. ContactDB uses a thermal camera to observe the thermal after-prints left by heat transfer from hand to object during grasping. It is thus able to texture the object mesh surface with high-resolution contact maps. Participants grasp the objects with one of two post-grasp functional intents: \textit{handoff} or \textit{use}. Since ContactDB contact maps have continuous values $t(p_i) \in [0, 1]$, we threshold them at $\tau_t$ to determine whether a point is attractive or repulsive. Figure~\ref{fig:contact_map} shows an example of human contact demonstration from the ContactDB dataset.
\begin{equation}\label{eq:contact_heatmap}
    c_i =
    \begin{cases}
        +1, & \text{if } t(p_i) \geq \tau_t\\
        -1, & \text{otherwise}
    \end{cases}
\end{equation}
\section{Hand Models}\label{sec:hand_models}
\begin{figure}[h!]
    	\centering
    	\includegraphics[width=0.48\textwidth]{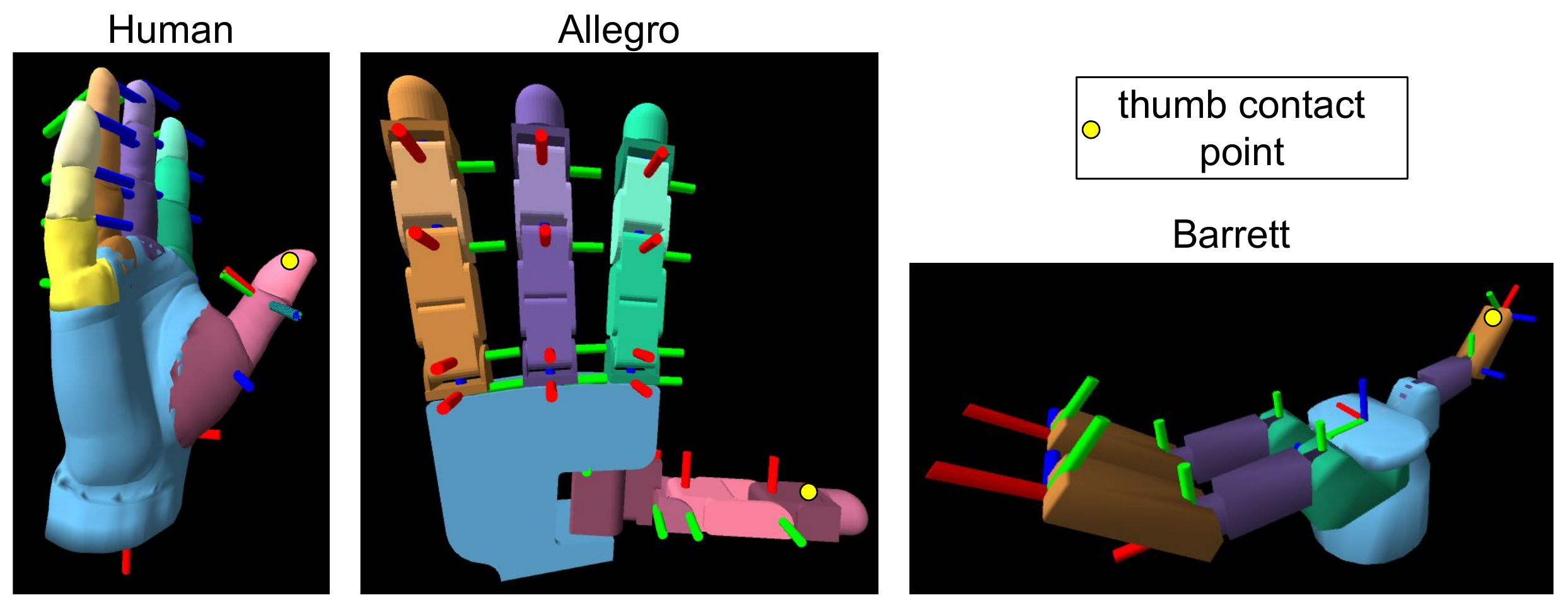}  	
    	\caption{Hand models used in our experiments, along with joint axes and special thumb contact point. Left: HumanHand~\cite{miller2004graspit}, middle: Allegro Hand~\cite{allegro_hand}, right: Barrett Hand~\cite{barrett_hand}}
    \label{fig:hand_models}
\end{figure}

Given the contact map on the object surface, an articulated hand model is used to set up an optimization problem which seeks a hand pose that agrees with the observed contact map. We use three kinematically diverse hand models to show that the object-centric contact representation of ContactGrasp allows grasp synthesis with diverse hand models (See Table~\ref{tab:hand_models} and Figure~\ref{fig:hand_models}). Each model is a kinematic tree, with parts modelled by rigid meshes. In addition to the articulation DOFs $d_i$ the overall position of the hand in 3D space is defined by a 6 DOF rigid body transform T. We denote the full pose of the hand as $\Phi = \left(T, \left\{d_i\right\}_{i=1}^{D}\right)$, where $D$ is the number of articulation DOFs of the hand. We compute a signed distance field (SDF) around each hand part and attach it to the part's local coordinate system to support hand pose optimization. See Section~\ref{sec:grasp_synthesis} for further details.

\begin{table}
\begin{tabular}{c|c|c|c}
\textbf{Model} & \textbf{Fingers} & \textbf{Joints} & \textbf{Articulation DOFs}\\
\hline
HumanHand~\cite{miller2004graspit} & 4 fingers, 1 thumb & 15 & 20\\
Allegro~\cite{allegro_hand} & 3 fingers, 1 thumb & 12 & 16\\
Barrett~\cite{barrett_hand} & 3 digits & 7 & 4\\
\end{tabular}
\caption{Hand models used in our experiments.}
\label{tab:hand_models}
\end{table}
\section{ContactGrasp: Grasp Synthesis} \label{sec:grasp_synthesis}
\begin{figure*}[ht!]
    \centering
    \includegraphics[width=0.95\textwidth]{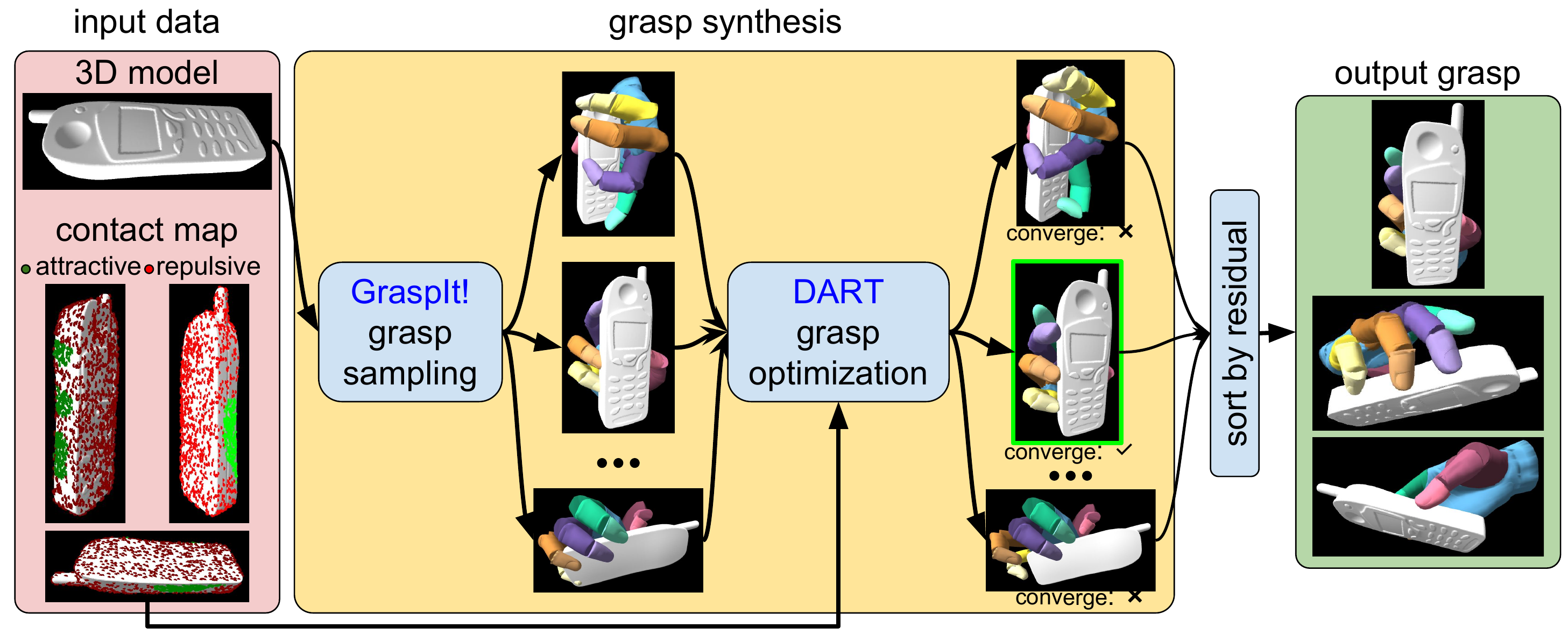}
    \caption{Overview of the ContactGrasp algorithm. GraspIt!~\cite{ciocarlie2007dimensionality} is used to sample random grasps for the object geometry. These are then refined and ranked for agreement with a human-demonstrated contact map to synthesize functional grasps.}
    \label{fig:overview}
\end{figure*}

In this section, we describe the algorithm to synthesize grasps from object geometry and a contact map. Grasp synthesis consists of estimating the full configuration $\Phi$ of the hand, which includes the 6-DOF `palm' pose as well as joint values. This is done through an appropriately initialized nonlinear optimization. Figure~\ref{fig:overview} shows an overview of the entire algorithm.

\subsection{Grasp Optimization}
\begin{figure}[h!]
    \centering
    \includegraphics[width=0.45\textwidth]{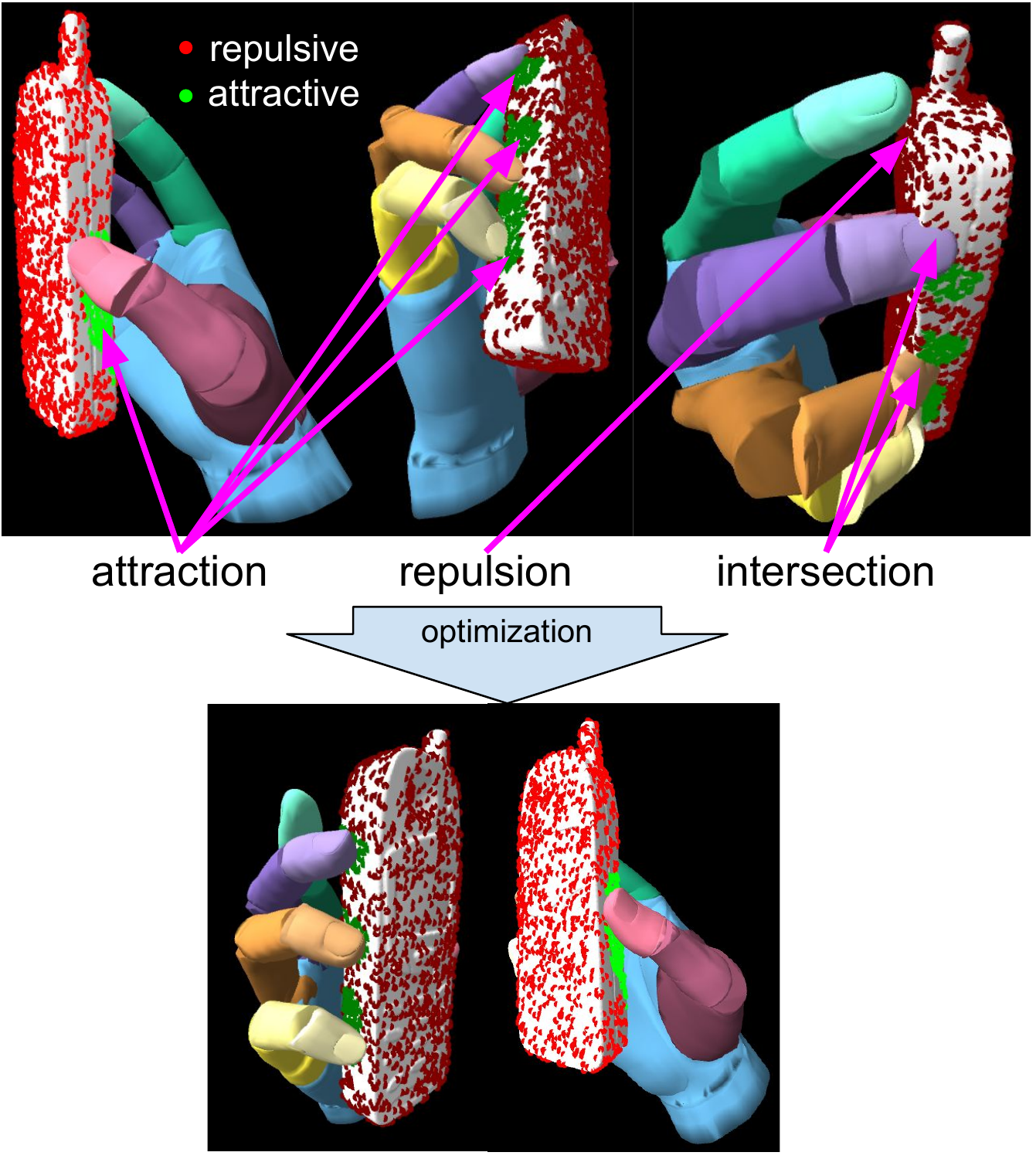}
    \caption{Top: Various factors involved grasp optimization. Bottom: Optimized result.}
    \label{fig:dart_optim}
\end{figure}

Intuitively, the objective of this optimization is to encourage contact at attractive points and discourage contact at repulsive points in the contact map. The full objective function consists of three terms, inspired from~\cite{schmidt2015dartp} (see Figure~\ref{fig:dart_optim} for a visual example of each term):

\subsubsection{Grasp Term}
\begin{figure}
    \centering
    \includegraphics[width=0.30\textwidth]{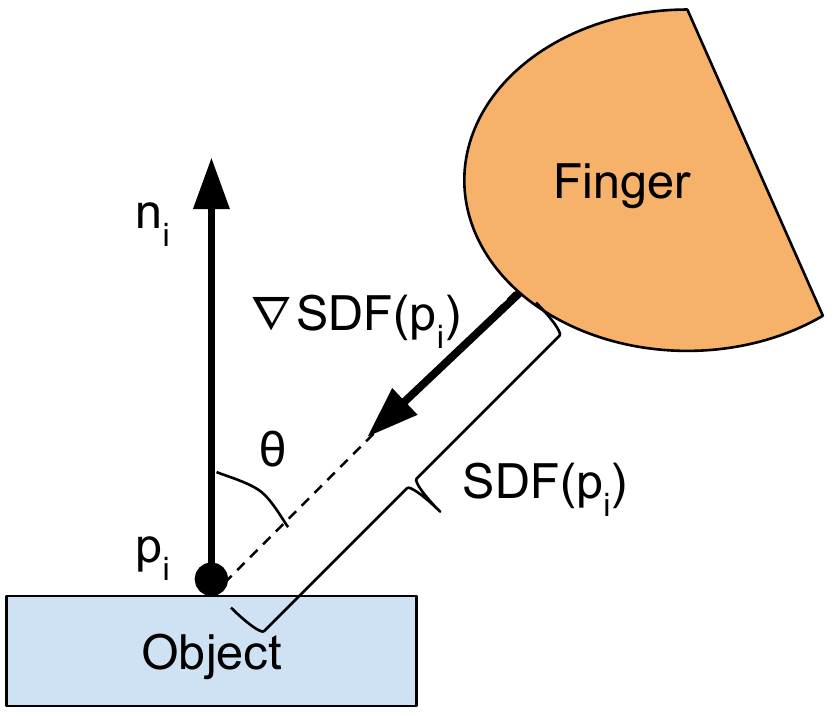}
    \caption{Geometry of the activation function for a repulsive point $p_i$ in the contact map for an object.}
    \label{fig:dart_repulsive_point}
\end{figure}

The grasp term attracts (resp. repels) the closest hand segment to every attractive (resp. repulsive) point in the contact map. For a given contact map $\mathbf{c}$ and hand pose $\Phi$, the term is stated as:
\begin{align}\label{eq:grasp_obj}
    L_{grasp}(\Phi | \mathbf{c}) &= \sum_{i=1}^N \lambda_a [c_i = +1] SDF_{k_i}(p_i)^2 \nonumber\\
    &- \lambda_r [c_i = -1] SDF_{k_i}(p_i)^2
\end{align}
Where $k_i$ is the index of the hand segment that is closest to point $p_i$, $SDF_k(\cdot)$ is the signed distance function (SDF) associated with hand segment $k$, and $[\cdot]$ is the indicator function. $\lambda_a$ and $\lambda_r$ are hyperparameters controlling the strength of attractive and repulsive points. However, Eq.~\ref{eq:grasp_obj} unnecessarily penalizes some hand poses. As shown in Figure~\ref{fig:dart_repulsive_point}, we want to repulse a hand part away from a repulsive point $p_i$ only if it is directly above the repulsive point i.e. the vector connecting $p_i$ to the nearest point on the hand is parallel to the surface normal at $p_i$. Since SDF measures Euclidean distances, it penalizes a nearby hand part even if it is not directly above $p_i$. Hence we modify the activation condition for repulsive points in Eq.~\ref{eq:grasp_obj} as follows, taking the surface normal $n_i$ at point $p_i$ into consideration (see also Figure~\ref{fig:dart_repulsive_point}):
\begin{align}\label{eq:grasp_obj_final}
    L_{grasp}(\Phi | \mathbf{c}) &= \sum_{i=1}^N \lambda_a [c_i = +1]SDF_{k_i}(p_i)^2 \nonumber\\
    &- \lambda_r [c_i = -1]f_{k_i}(p_i, n_i)
\end{align}
where
\begin{equation}\label{eq:cylindrical_repulsion}
    f_{k_i}(p_i, n_i) =
    \begin{cases}
    SDF_{k_i}(p_i)^2, & \text{if } |\widehat{\nabla SDF}_{k_i}(p_i) \cdot n_i| > \tau_n\\
    0, & \text{otherwise}
    \end{cases}
\end{equation}
where $\widehat{\nabla SDF}(\cdot)$ is the unit vector in the direction of the gradient of the SDF.

\subsubsection{Thumb Contact Term}
It is well known that the thumb is especially important in human grasps~\cite{cutkosky1990human}. To account for this, we specify a point $p_{thumb}$ on the thumb (or the part closest resembling the thumb) in the hand model (yellow dots in Figure~\ref{fig:hand_models}). The thumb contact term encourages $p_{thumb}$ to be in contact with the object:
\begin{equation}
L_{thumb}(\Phi) = \lambda_t SDF_{object}(p_{thumb})^2
\end{equation}
where $SDF_{object}(\cdot)$ is the signed distance function associated with the object, and hyperparameter $\lambda_t$ controls the strength of this term.

\subsubsection{Intersection Term}
The intersection term $L_{int}(\Phi)$ discourages intersection of the hand model with the object and self-intersection among segments of the hand, and is the same as the one used in~\cite{schmidt2015dartp}. Its strength is controlled by the hyperparameter $\lambda_i$.

\subsubsection{Optimization}
The full objective function for grasp optimization is given by
\begin{equation}\label{eq:final_obj}
    L(\Phi | \mathbf{c}) = L_{grasp}(\Phi | \mathbf{c}) + L_{thumb}(\Phi) + L_{int}(\Phi)
\end{equation}
We use Dense Articulated Real-time Tracking (DART)~\cite{schmidt2014dart} to minimize Eq.~\ref{eq:final_obj} and get the optimized hand pose. Specifically, we modify the Contact Prior mechanism in DART~\cite{schmidt2015dartp} to 1) support repulsive points, and 2) remove the depth-map observation term. DART approximately minimizes Eq.~\ref{eq:final_obj} by running the Levenberg-Marquadt algorithm (see~\cite{schmidt2014dart} for more details).

\subsection{Initializing the Grasp Optimization}
Since the Levenberg-Marquadt grasp optimization is local, and search in the high-dimensional hand pose space has many local minima, providing good initialization to the optimizer becomes important. Towards this end, we develop an algorithm to sample diverse grasps for the object geometry which are agnostic to the contact map, using the publicly available GraspIt! grasp planner~\cite{miller2004graspit,ciocarlie2007dimensionality}. These are later ranked for agreement with the contact map using the residue after minimizing Eq~\ref{eq:final_obj}.

\begin{figure*}[h!]
\centering
\includegraphics[width=0.95\textwidth]{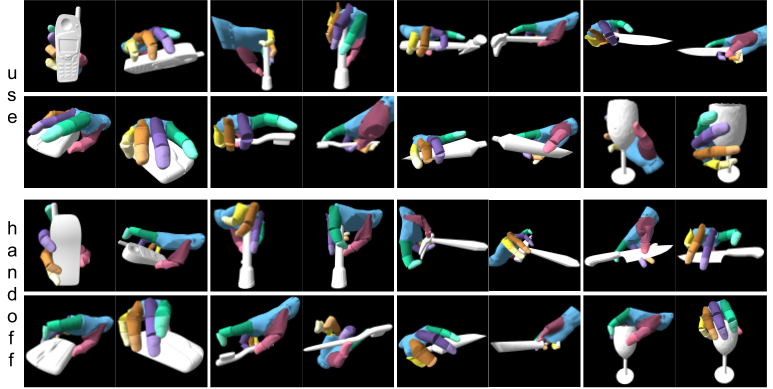}
\caption{Functional HumanHand grasps synthesized by ContactGrasp. Top: Use, bottom: Hand-off.}
\label{fig:human_grasps}
\end{figure*}

\begin{figure*}[h!]
\centering
\includegraphics[width=0.95\textwidth]{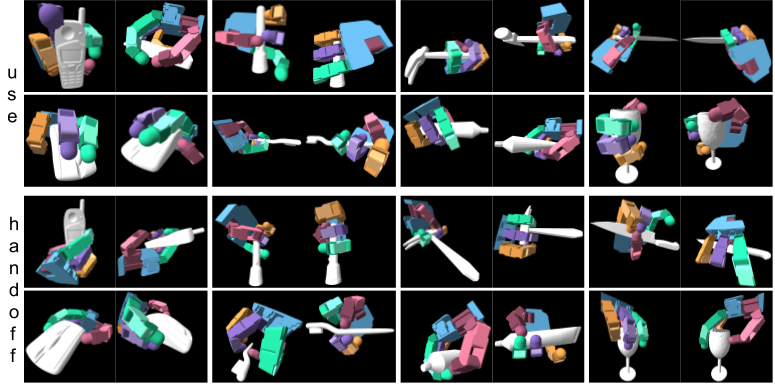}
\caption{Functional Allegro hand grasps synthesized by ContactGrasp. Top: Use, bottom: Hand-off.}
\label{fig:allegro_grasps}
\end{figure*}

\begin{figure*}[h!]
\centering
\includegraphics[width=0.95\textwidth]{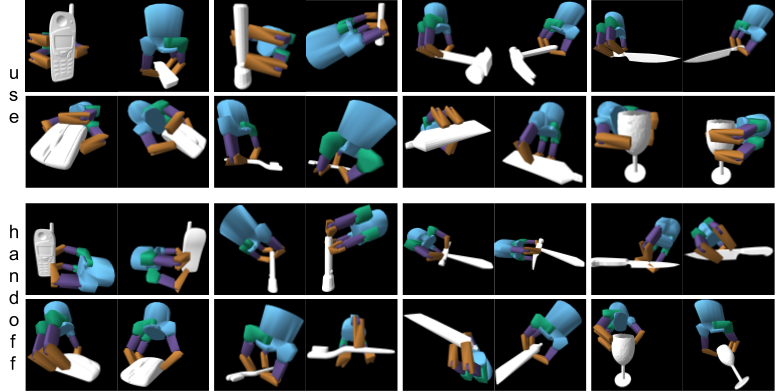}
\caption{Functional Barrett hand grasps synthesized by ContactGrasp. Top: Use, bottom: Hand-off.}
\label{fig:barrett_grasps}
\end{figure*}

\begin{figure*}[h!]
\centering
\includegraphics[width=0.95\textwidth]{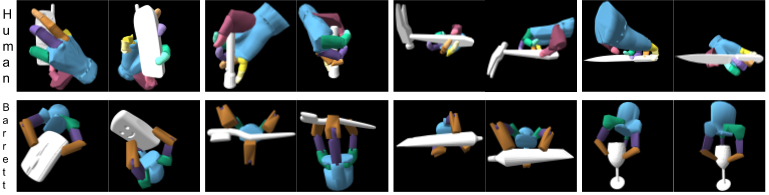}
\caption{Top-ranked grasps from GraspIt!~\cite{ciocarlie2007dimensionality}, which are agnostic to contact map and hence are not functional, especially for the `use' intent. For example, flashlight button is not accessible by the thumb, finger touches knife blade, cellphone screen and wineglass opening are blocked by palm. Top: Human Hand, Bottom: Barrett Hand.}
\label{fig:graspit_grasps}
\end{figure*}

Specifically, we seed the planner with a coarse grasp $\phi = (\mathbf{a}, \theta, d)$, where $\mathbf{a}$ is the approach point on the object surface, $\theta$ is the roll angle around approach vector and $d$ is the distance from object surface. We use the negated surface normal $-\mathbf{n}$ at the approach point $\mathbf{a}$ as the approach vector, making since the hand approach anti-parallel to the object surface normal (similar to~\cite{pelossof2004svm}). Approach points $\mathbf{a}$ are sampled uniformly at random over the object surface, whereas $\theta$ and $d$ are set from the discrete sets \{0, 90, 180, 270\} and \{0 cm, 1 cm, 2 cm, 3 cm\} respectively. GraspIt!’s Simulated Annealing planner then runs for 45K iterations for each seed to optimize the Contact Energy cost function~\cite{ciocarlie2007dimensionality}, exploring in a cone around the specified approach vector. We pick the top 2 grasps after each run of the planner, and add them to the set of sampled diverse grasps $\mathcal{D}$. Note that $\mathcal{D}$ contains full grasps, since GraspIt! provides a bridge to convert coarse grasps $\phi$ to full grasps $\Phi$ through its planner. Figure~\ref{fig:overview} shows some grasps sampled in this manner for a cellphone.

The last step is to refine and rank the grasps in $\mathcal{D}$ by running grasp optimization on each of them, and considering the residual after convergence (Eq.~\ref{eq:final_obj}) as the negative score. Most of the grasps in $\mathcal{D}$ result in large residuals because they are agnostic to the contact maps and hence can be out of the basin of convergence of the local optimization. Hence this ranking process causes the ‘correct’ hand pose to show up among the top-ranked, from which it can be easily identified.

To summarize, the process described in this section is recovers the correct full hand pose that agrees with a given contact map on an object. Manually annotating the full hand pose prohibitively expensive because of the high dimensionality.
\section{Results} \label{sec:results}

We use the ContactGrasp algorithm (Section~\ref{sec:grasp_synthesis}) to synthesize grasps for a diverse set of household objects for two different post-grasp functional intents: \textit{using} the object, and \textit{handing it off}. Grasps are synthesized for the three hand models described in Section~\ref{sec:hand_models}. We use human contact demonstrations for functional grasps from the ContactDB dataset~\cite{contactdb}. 25 objects in ContactDB have demonstrations for both using the object and handing it off. From these, we select a subset of 19 objects (see supplementary video for a list) for which bi-manual grasps were not observed. Note that the grasp optimization strategy described in Section~\ref{sec:grasp_synthesis} supports bi-manual grasps. However, we focus on single-handed grasps in this paper owing to lack of left-handed hand models and need for a more complex initialization strategy for the optimization.

We set the hyperparameter $\tau_t$ (threshold on contact map value) to 0.3, $\lambda_a$ (attractive contact point strength) to 150.0, $\lambda_r$ (repulsive contact point strength) to 20.0, $\lambda_t$ (thumb contact point strength) to 25.0, and $\lambda_i$ (intersection term strength) to 100.0.

\subsection{Qualitative results}
Figures~\ref{fig:human_grasps}, ~\ref{fig:allegro_grasps}, and~\ref{fig:barrett_grasps} show the synthesized grasps for the HumanHand, Allegro hand and Barrett hand respectively, for 8 objects and 2 functional intents(see supplementary video for other objects). They demonstrate functional grasps e.g. for the `use' intent, the flashlight button is easily accessible by the thumb, fingers rest on mouse click buttons, knife is held by the handle. In contrast, the top-ranked grasps from GraspIt! are shown in Figure~\ref{fig:graspit_grasps}. They are stable but do not demonstrate functionality, especially for the `use' intent e.g. flashlight button is not accessible by the thumb, finger touches knife blade, cellphone screen and wineglass opening are blocked by palm.

\begin{table}[h!]
\centering
\begin{tabular}{c|c|c|c}
\multirow{2}{*}{\textbf{Intent}} & \multirow{2}{*}{\textbf{End-effector}} & \textbf{ContactGrasp} & \textbf{GraspIt!}\\
&& $L_{grasp}$ &  $L_{grasp}$\\
\hline
\multirow{4}{*}{use} & HumanHand~\cite{miller2004graspit} & \textbf{-0.07} & 0.17\\
& Allegro~\cite{allegro_hand} & -0.06 & 0.32\\
& Human - Allegro~\cite{tosun2014general} & 0.08 & 0.12\\
& Barrett~\cite{barrett_hand} & -0.03 & 0.19\\
\hline
\multirow{4}{*}{handoff} & HumanHand~\cite{miller2004graspit} & \textbf{-0.14} &  0.19\\
& Allegro~\cite{allegro_hand} & -0.11 & 0.41\\
& Human - Allegro~\cite{tosun2014general} & 0.05 & 0.15\\
& Barrett~\cite{barrett_hand} & -0.04 & 0.22\\
\end{tabular}
\caption{\textbf{Dis}agreement of the ContactGrasp and GraspIt! grasps from human-demonstrated contact, measured by $L_{grasp}$ (see Eq.~\ref{eq:grasp_obj_final}, lower is better). Human-Allegro indicates grasps mapped from human to the Allegro model using~\cite{tosun2014general}.}
\label{tab:grasp_residue}
\end{table}

\subsection{Quantitative results}
Table~\ref{tab:rank_analysis} shows the median rank (across all objects) of the synthesized grasp, when grasps are ranked according to 1) the DART residual, and 2) GraspIt!'s Contact Energy metric~\cite{ciocarlie2007dimensionality}. The median rank is significantly lower for the former. This indicates that approaches like~\cite{ciocarlie2007dimensionality} that consider only object geometry cannot guarantee functional contact. ContactGrasp can effectively leverage these approaches to synthesize functional grasps using demonstrations.

\begin{table}
\centering
\begin{tabular}{c|c|c|c}
\textbf{Intent} & \textbf{End-effector} & \textbf{ContactGrasp rank} & \textbf{GraspIt! rank}\\
\hline
\multirow{3}{*}{use} & HumanHand~\cite{miller2004graspit} & \textbf{3(0.03\%)} &  4484(34.41\%)\\
& Allegro~\cite{allegro_hand} & 22(0.69\%) & 1289(40.28\%)\\
& Barrett~\cite{barrett_hand} & 6(0.19\%) & 2362(74.09\%)\\
\hline
\multirow{3}{*}{handoff} & HumanHand~\cite{miller2004graspit} & \textbf{1(0.01\%)} &  3751(38.37\%)\\
& Allegro~\cite{allegro_hand} & 22(0.56\%) & 1258(38.20\%)\\
& Barrett~\cite{barrett_hand} & 24(0.75\%) & 830(25.94\%)\\
\end{tabular}
\caption{Median rank of the correct grasp (lower is better).}
\label{tab:rank_analysis}
\end{table}

Table~\ref{tab:grasp_residue} shows further quantitative results in the form of the $L_{grasp}$ value (see Eq.~\ref{eq:grasp_obj_final}) for the synthesized grasp as well as the top-ranked grasp by GraspIt!'s Contact Energy metric. $L_{grasp}$ measures the grasp's disagreement with the demonstrated contact map, and is significantly lower for ContactGrasp grasps. This verifies that ContactGrasp produces grasps that are significantly closer to the human demonstrations than GraspIt!, which is agnostic to the demonstrations.

In addition, we implement the kinematic retargeting algorithm of Tosun et at~\cite{tosun2014general} for mapping the synthesized human hand grasp to the Allegro model. Each finger is treated as a separate kinematic chain (sampled with 50 points) to be re-targeted. The human pinky finger is discarded. The $L_{grasp}$ values for these mapped grasps are higher than those for the ContactGrasp Allegro grasps as well as human grasps. This shows that deterministic mapping of grasps between hand models does not reliably reproduce contact, supporting our motivation for developing ContactGrasp.

\subsection{Failure Cases}
\begin{figure}
\centering
\includegraphics[width=0.4\textwidth]{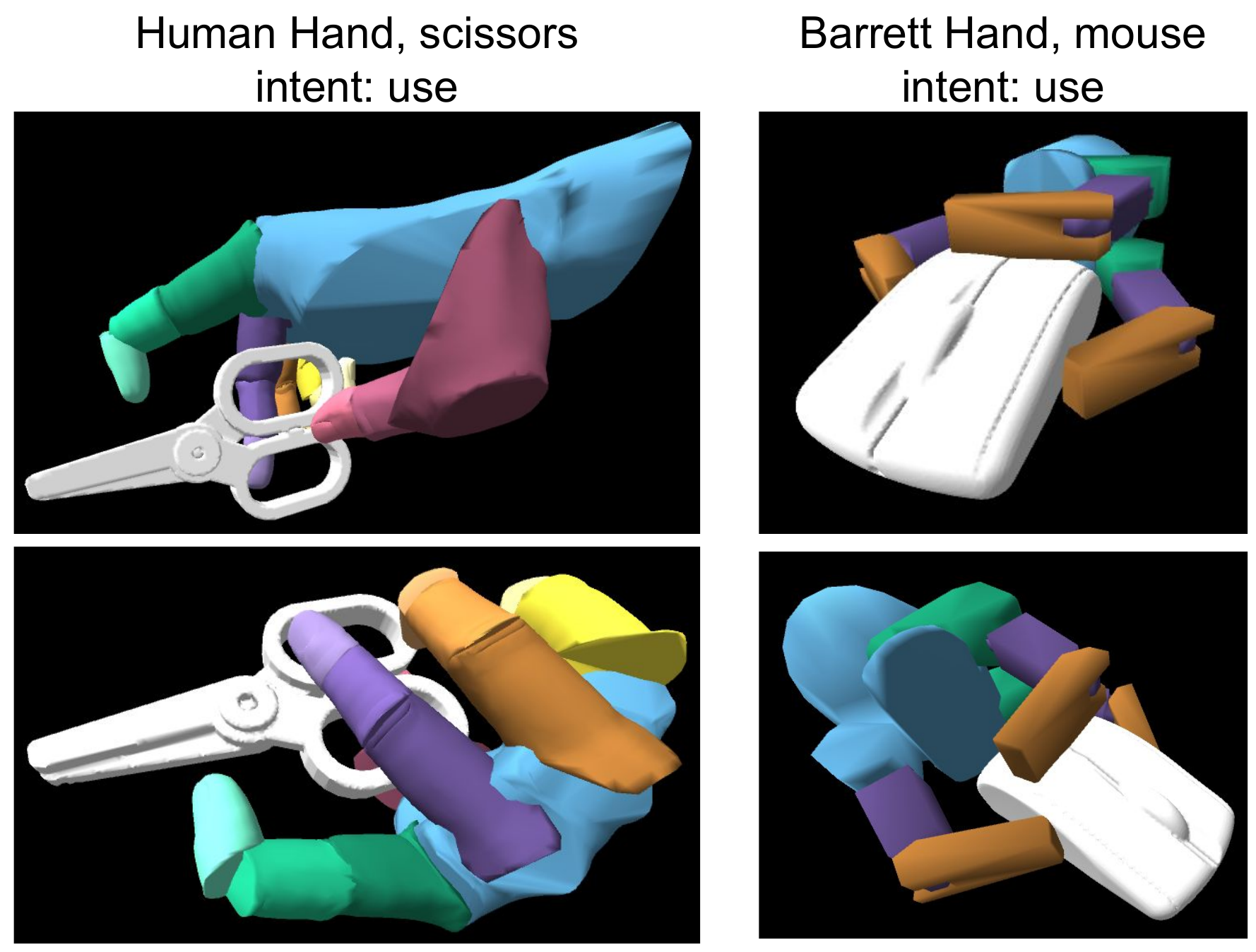}
\caption{Failure cases. Left: Grasp sampling is unlikely to produce a grasp with fingers in narrow holes, and optimization then gets stuck in local minima. Right: Some end effectors lack the structure for functional grasps e.g. resting fingers on mouse buttons.}
\label{fig:failure_cases}
\end{figure}

Figure~\ref{fig:failure_cases} shows failure cases of ContactGrasp. These occur when the grasp sampling stage is not exhaustive enough to sample fine manipulation behaviors like fingers through holes, or when the target hand model does not possess the geometry to be able to achieve a complicated control pattern.

\section{Conclusion}
To summarize, we develop a multi-point contact model in this paper, which can plug-and-play with kinematically diverse hand models to synthesize grasps. These grasps can be modulated by hand-object contact demonstrations to be functional, supporting post-grasp actions like using the object or handing it off. We show that our approach ContactGrasp, which directly optimizes for contact, is superior than other approaches which kinematically re-target observed human grasps to the target hand model. We demonstrate the effectiveness of ContactGrasp by synthesizing functional grasps for 3 significantly diverse hand models, 19 household objects, and 2 functional intents.

\bibliographystyle{IEEEtran}
\bibliography{IEEEabrv,references}

\end{document}